\documentclass{article}
\usepackage{ijcai24}

\usepackage{times}
\usepackage{soul}
\usepackage{url}
\usepackage[hidelinks]{hyperref}
\usepackage[utf8]{inputenc}
\usepackage[small]{caption}
\usepackage{graphicx}
\usepackage{amsmath}
\usepackage{bbding}
\usepackage{amsthm}
\usepackage{amsfonts}
\usepackage{booktabs}
\usepackage{algorithm}
\usepackage{algorithmic}
\usepackage{multirow}
\usepackage[switch]{lineno}
\usepackage{xcolor}
\usepackage{xspace}
\usepackage{subfigure}
\usepackage{array}
\usepackage{tikz}
\usepackage{bm}
\usepackage{tikzsymbols}

\makeatletter
\DeclareRobustCommand\onedot{\futurelet\@let@token\@onedot}
\def\@onedot{\ifx\@let@token.\else.\null\fi\xspace}

\def\eg{\emph{e.g\onedot}} 
\def\ie{\emph{i.e\onedot}} 

\makeatother

\title{An Embarrassingly Simple Approach to Enhance Transformer Performance in Genomic Selection for Crop Breeding}

\author{
Renqi Chen$^{1}$
\and
Wenwei Han$^{1,2,3}$\and
Haohao Zhang$^{4}$\and
Haoyang Su$^{1}$\and
Zhefan Wang$^{1}$\and\\
Xiaolei Liu$^{5}$\and
Hao Jiang$^{2,3}$\and
Wanli Ouyang$^{1}$\And
Nanqing Dong$^1$
\affiliations
$^1$Shanghai Artificial Intelligence Laboratory\\
$^2$Institute of Computing Technology, Chinese Academy of Sciences\\
$^3$School of Computer Science, University of Chinese Academy of Sciences\\
$^4$School of Computer Science and Artificial Intelligence, Wuhan University of Technology\\
$^5$School of Animal Science and Technology, Huazhong Agricultural University
\emails
\{chenrenqi, hanwenwei, dongnanqing\}@pjlab.org.cn\\
}

\begin{document}    

\maketitle

\begin{abstract}
Genomic selection (GS), as a critical crop breeding strategy, plays a key role in enhancing food production and addressing the global hunger crisis. The predominant approaches in GS currently revolve around employing statistical methods for prediction. However, statistical methods often come with two main limitations: strong statistical priors and linear assumptions. A recent trend is to capture the non-linear relationships between markers by deep learning. However, as crop datasets are commonly long sequences with limited samples, the robustness of deep learning models, especially Transformers, remains a challenge. In this work, to unleash the unexplored potential of attention mechanism for the task of interest, we propose a simple yet effective Transformer-based framework that enables end-to-end training of the whole sequence. Via experiments on \textit{rice3k} and \textit{wheat3k} datasets, we show that, with simple tricks such as k-mer tokenization and random masking, Transformer can achieve overall superior performance against seminal methods on GS tasks of interest.

\end{abstract}

\section{Introduction}
Grain production security serves as the cornerstone of human existence, exerting a pivotal role in shaping the health, stability, and prosperity of global society. Addressing global hunger not only aligns with the United Nations Sustainable Development Goals (SDGs)~\cite{17goals} of Zero Hunger, but also adheres to the Leaving No One Behind Principle (LNOB)~\cite{LNOB}. However, according to the report by the Food and Agriculture Organization of the United Nations in 2023~\cite{fao2023state}, around 735 million people worldwide were suffering from hunger. Therefore, utilizing technological advancements to boost grain production is vital for eliminating hunger by 2030. Crop breeding stands as a fundamental approach to enhancing grain production. Continuously improving crop varieties to increase yield and resilience effectively boosts grain production, thereby meeting the escalating demand for food.

\begin{table}[t]
\centering
\resizebox{\columnwidth}{!}{
\begin{tabular}{c|cc|cc}\toprule
\multirow{2}{*}{{Method}} & \multicolumn{2}{c|}{Assumption} & \multicolumn{2}{c}{Training} \\ \cmidrule{2-5}
 & {Stats} & {Linear} & {End2End} & {GPU}\\ 
\midrule
GBLUP & \multirow{2}{*}{{\tikz{\draw[thick] (0,0) circle (4pt);}}} & \multirow{2}{*}{{\tikz\draw[thick] (0,0) circle (4pt);}} & \multirow{2}{*}{} & \multirow{2}{*}{}\\
\cite{vanraden2008efficient} &  &  &  & \\
BayesA & \multirow{2}{*}{{\tikz\draw[thick] (0,0) circle (4pt);}} & \multirow{2}{*}{{\tikz\draw[thick] (0,0) circle (4pt);}} & \multirow{2}{*}{} & \multirow{2}{*}{}\\
\cite{meuwissen2001prediction} &  &  &  & \\
DLGWAS & \multirow{2}{*}{{\tikz\draw[thick] (0,0) circle (4pt);}} & \multirow{2}{*}{{\tikz\draw[thick] (0,0) circle (4pt);}} & \multirow{2}{*}{} & \multirow{2}{*}{\Checkmark}\\
\cite{manor2013predicting} &  &  &  & \\
DNNGP & \multirow{2}{*}{} & \multirow{2}{*}{{\tikz\draw[thick] (0,0) circle (4pt);}} & \multirow{2}{*}{} & \multirow{2}{*}{\Checkmark}\\
\cite{wang2023dnngp} &  &  &  & \\ \midrule
Ours & {} & {} & {\Checkmark} & {\Checkmark}\\
\bottomrule
\end{tabular}
}
\caption{Comparison with the state-of-the-art methods for genotype-to-phenotype prediction task. ``Stats'' and ``Linear'' denote whether the statistical prior and linear assumption is required for the method, respectively. ``End2End'' and ``GPU'' denote whether the model can be trained in an end-to-end fashion and supported by GPU. ``{\Large{$\circ$}}" and ``$\checkmark$" denote conformity.}
\label{tab:baseline_compare}
\end{table}

Genomic selection (GS), proposed by~\cite{meuwissen2001prediction}, is regarded as a promising breeding paradigm~\cite{boichard2012genomic,garcia2016changes}. GS usually involves predicting the phenotypes of polygenic traits in plants or crops by using high-density markers covering the entire genome. By implementing early screening of candidate populations, the genetic process can be accelerated, reducing generation intervals and significantly shortening breeding cycles. Unlike phenotype selection~\cite{siepielski2013spatial,kingsolver2007patterns} and marker-assisted selection~\cite{ribaut1998marker,xu2008marker}, GS can utilize single nucleotide polymorphism (SNP) data obtained from organisms. SNP sequences are \textit{loci} sequences with two or more alleles extracted from DNA sequences, which are the most widely adopted genetic markers in GS.\footnote{Stripping away segments that exhibit the same patterns from DNA sequences to obtain SNP sequences is advantageous for analyzing genotype-phenotype associations, as phenotypic differences often arise from distinct segments between two DNA sequences.} 

A mainstream solution to GS is to mine the statistical knowledge of SNP data. GBLUP~\cite{vanraden2008efficient} and SSBLUP~\cite{goddard2011using} utilize kinship matrices to weight and aggregate the phenotypic values of different individuals, thereby obtaining their estimated breeding values (EBVs) but may not fully explore and utilize genomic information. Another category of statistical methods involves estimating marker effects on a reference population and then predicting marker effects on a candidate population, followed by linear aggregation to obtain EBVs for the candidates~\cite{meuwissen2001prediction,tipping2003fast,de2009predicting,karkkainen2012back}. These methods typically demand substantial computational resources and lack parallelization capabilities. Their practicality in time-sensitive breeding scenarios is constrained. In addition, two common limitations of statistical methods are the strong statistical priors (\eg, Gaussian distribution) and linear relationship assumption between markers. Though these two assumptions simplify the statistical computation, statistical models suffer a setback when the underlying data distribution differs.   

In order to capture the non-linear relationships between markers, efforts have been made by applying deep learning to process SNP sequences, especially with convolutional neural networks (CNN)~\cite{ma2017deepgs,liu2019phenotype,xie2023residual} and Transformers~\cite{wu2024transformer}. However, two major data challenges threaten the robustness of deep learning models. First, the number of samples for each crop species is limited. This can easily cause overfitting for deep learning models, while statistical models are commonly deemed as robust solutions. Second, SNP are long sequences. Besides, plant genomes, unlike those of animals, demonstrate a higher frequency of long insertions and deletions attributed to the activity of transposable elements~\cite{ramakrishnan2022transposable}, leading to a greater abundance of SNP data among different plant genomes. The locality of convolution introduces a strong inductive bias for local dependencies, thus cannot catch the long-range interactions between markers. On the contrary, while Transformers can better tackle long-range dependencies in sequences, attention's quadratic complexity becomes the major bottleneck for long sequence~\cite{dao2022flashattention}. 

To mitigate the effect of long sequences, an intuitive way is to statistically pre-process the data before building deep learning models. However, the limitations of statistical methods still remain. For example, DNNGP utilizes principal component analysis (PCA), a statistical dimensionality reduction technique, and condenses the sequence to only several hundred dimensions for CNN modeling~\cite{wang2023dnngp}. However, the viability of this hinges on fulfilling the assumption of PCA which states that there exist linear correlations among variables in the dataset, but this might not be true for the majority of SNP datasets. Thus, the linear dimensionality reduction techniques may result in information loss regarding \emph{loci} that influences phenotypes. Another statistical way to shorten the sequence is traditional feature selection, \ie, selecting important features out of hundreds of thousands of SNPs~\cite{manor2013predicting}. For example, a promising method is to employ genome-wide association studies (GWAS) to identify major effect loci within gene sequences~\cite{liu2019phenotype}. However, these methods heavily rely on the accuracy of GWAS and are not conducive to modeling and analyzing quantitative traits. 

To address the aforementioned challenges of statistical and deep learning methods, we propose a simple yet effective Transformer-based method that supports end-to-end training. The proposed method leverages simple tricks such as k-mer tokenization, and random masking to achieve robust predictive performance from genotypes to phenotypes on crop breeding datasets. It is worth mentioning that, though similar natural language processing (NLP) techniques have been investigated on DNA data due to the similarities between DNA sequences and text sequences~\cite{ji2021dnabert}, there is no such application on SNP data yet. \textbf{To the best of our knowledge, we are the first to successfully adapt these simple NLP techniques to address the GS problem.}
Compared with statistical methods, our method does not require rigid statistical priors or linear assumptions, and thus can better capture the non-linear relationships. Supported by GPU, our method has a significantly shorter inference time. Compared with existing deep learning methods, our method can better leverage the attention mechanism~\cite{vaswani2017attention} to assist contextual understanding. 
We summarize the differences between our method and seminal methods in Tab.~\ref{tab:baseline_compare}. 

We extensively evaluate the robustness of our method on two crop datasets, \textit{rice3k}~\cite{wang2018genomic} and \textit{wheat3k}. \textit{rice3k} is a public dataset on rice with SNP data. On the \textit{rice3k} dataset, we outperform the current best-performing method (a hybrid method of GWAS and Transformer) on average over $1.05\%$ in the accuracy metric. \textit{wheat3k} is a private cereal grain dataset to be released, which contains 3032 SNP sequences following a similar setup of \textit{rice3k}. Our method also consistently outperforms the seminal baselines on \textit{wheat3k}.

\begin{itemize}
    \item We propose an end-to-end Transformer-based framework for genomic selection that enables capturing non-linear relationships between genotypes and phenotypes.
    \item We show that in genotype-to-phenotype prediction tasks, using k-mer tokenization and random masking can effectively reduce the data complexity while enhancing model prediction performance.
    \item We conduct extensive experiments on the \textit{rice3k }and \textit{wheat3k} datasets, where our method achieves the state-of-the-art performance on both datasets against mainstream methods.
\end{itemize}

\section{Related Work}
\subsection{Analysis-based Genomic Selection}
Currently, analysis-based genomic selection methods are mainly classified into two families: \emph{best linear unbiased prediction} (BLUP) methods and Bayesian methods \cite{gualdron2020performances}. BLUP-based methods evaluate random effects using a linear model. GBULP~\cite{vanraden2008efficient} constructs a pedigree relationship matrix using genomic information, incorporating it as a random effect into the model to estimate genetic values or predict trait values of individuals. RRBLUP~\cite{endelman2011ridge} builds upon GBLUP by incorporating the concept of ridge regression. SSBLUP~\cite{goddard2011using} integrates both the genomic relationship matrix and pedigree relationship matrix, along with phenotypic data, into a unified mixed model. For Bayesian methods, such as~\cite{tipping2003fast,de2009predicting,karkkainen2012back}, the marker effects are assumed to follow different Gaussian distributions. For example, BayesA~\cite{meuwissen2001prediction} assumes that each marker has its own distribution and variance. Though Bayesian methods can reveal the relationship between genotype and phenotype to some extent, they are constrained by strong statistical priors and linear assumptions. In addition to the two families of methods, statistical learning methods~\cite{ke2017lightgbm} may also improve prediction accuracy on specific datasets. However, this improvement might not generalize, especially when the number of samples is significantly smaller than the number of feature dimensions.

\subsection{Deep Learning-based Genomic Selection}
Fueled by the recent success of deep learning in vision and language tasks, researchers have attempted to integrate deep learning into the Genome Selection (GS) field. DeepGS~\cite{ma2017deepgs} proposes a genome-wide selection framework based on CNN, while DualCNN~\cite{liu2019phenotype} utilizes a dual-stream CNN to predict quantitative traits. ResGS~\cite{xie2023residual} uses ResNet~\cite{he2016deep} to extract gene sequence information without using pooling layers. DNNGP~\cite{wang2023dnngp}, a seminal work, performs PCA to reduce the high dimensionality of gene data to extract effective information. However, the limited convolutional receptive field hinders the model's ability to capture linkage disequilibrium loci in long SNP sequences.

Transformer~\cite{vaswani2017attention} has exhibited excellent performance in modeling global interactions within sequences. GPformer~\cite{wu2024transformer} applies Informer~\cite{zhou2021informer}, a model from the long-term time series prediction field, to aggregate periodic information and capture region-specific features in SNP sequences. However, the computational complexity of Transformer also has a quadratic relationship with the sequence length. The current Transformer-based methods~\cite{wu2024transformer} are unable to handle the whole genetic SNP sequences for most crop species. There are hybrid methods~\cite{de2009predicting,liu2019phenotype} that use GWAS to pre-process the raw SNP sequence, but the results heavily depend on the results of GWAS and exhibit unstable performance across species. Instead, we propose to leverage tokenization and optimization acceleration techniques to achieve full attention computation on long SNP sequences, thereby mining long-range interactions and obtaining more accurate and robust predictive models.

\begin{table}[t]
\centering
\resizebox{\linewidth}{!}{
\begin{tabular}{c| c c c c c c c c c c c}
\toprule
\texttt{LETTER} & \texttt{A} & \texttt{T} & \texttt{C} & \texttt{G} & \texttt{Y} & \texttt{K} & \texttt{W} & \texttt{R} & \texttt{S} & \texttt{M} & \texttt{N} \\
\midrule
\textbf{A} & 2 & 0 & 0 & 0 & 0 & 0 & 1 & 1 & 0 & 1 & {-} \\ 
\textbf{T} & 0 & 2 & 0 & 0 & 1 & 1 & 1 & 0 & 0 & 0 & {-} \\ 
\textbf{C} & 0 & 0 & 2 & 0 & 1 & 0 & 0 & 0 & 1 & 1 & {-} \\
\textbf{G} & 0 & 0 & 0 & 2 & 0 & 1 & 0 & 1 & 1 & 0 & {-} \\
\textbf{N} & 0 & 0 & 0 & 0 & 0 & 0 & 0 & 0 & 0 & 0 & 1/2 \\
\bottomrule
\end{tabular}
}
\caption{Coding rule for SNP data. \textbf{A}, \textbf{T}, \textbf{C}, and \textbf{G} represent four basic nucleotides, while \textbf{N} represents an undetermined nucleotide that was not sequenced. Given a reference DNA sequence and a sample DNA sequence, at each nucleotide position, we use a set of \texttt{LETTER}s to represent possible combinations of two nucleotides. The numbers represent the frequency of occurrences. For example, \texttt{A} represents the combination \textbf{AA} and \texttt{W} represents the combination \textbf{AT} (\textbf{AT} and \textbf{TA} are equivalent). \texttt{N} means there is at least one \textbf{N}.}

\label{tab:def}
\end{table}

\subsection{Sequence Representation}
A straightforward method to handle SNP data is to directly model loci using additive encoding, which is widely adopted by seminal methods~\cite{vanraden2008efficient,aguilar2010hot,mittag2015influence}. But, it is overly rigid to represent genotypes by the number of non-reference alleles.

An emerging research topic is to leverage language models to model DNA data. However, unlike human language, where each word carries rich meaning, DNA's lexical units consist of only four basic nucleotide bases (\textbf{A}, \textbf{T}, \textbf{C}, \textbf{G}) with relatively ambiguous meanings, making them lower-level abstractions. For example, DNABERT~\cite{ji2021dnabert} and DNAGPT~\cite{zhang2023dnagpt} have shown that encoding DNA sequences into k-mer patterns can improve the performance on downstream tasks without losing information. DNABERT-2~\cite{zhou2023dnabert} uses random masking for unsupervised pre-training on DNA sequences. However, as there are fundamental differences between DNA and SNP sequences, it is still unclear whether these NLP techniques work on SNP data. In this work, we present the first empirical study.

\begin{figure*}[th]
  \centering
  \includegraphics[width=0.94\linewidth]{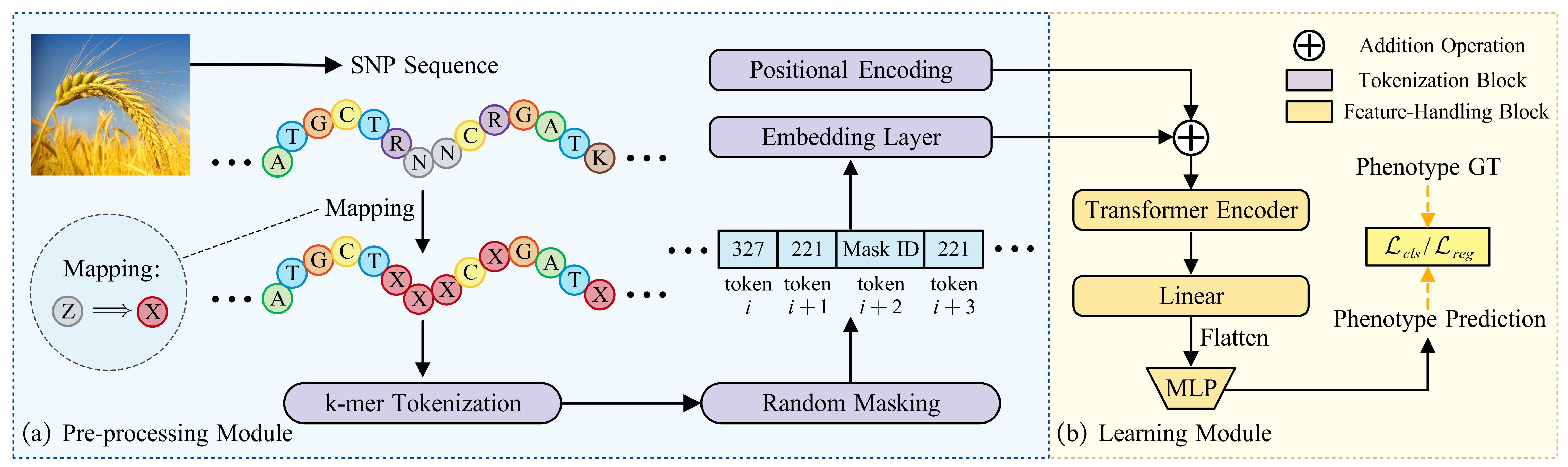}
  \caption{Illustration of the proposed framework, consisting of two modules, (a) a pre-processing module and (b) a learning module. (a) The raw SNP sequence is first pre-processed by a many-to-one mapping rule. Let $Z$ denote an arbitrary index that does not belong to the set, \ie, \{\texttt{A}, \texttt{T}, \texttt{C}, \texttt{G}\}, which are the four most frequent letters in SNP sequence. $Z$ is mapped to $X$ to reduce computational cost while it does not influence performance. The pre-processed sequence is then input into the tokenizer composed of k-mer and random masking to get token ID sequence and then the embedding layer. (b) Transformer encoder and MLP layers are adopted on embedding vectors to predict phenotype.}
  \label{fig:System Model}
\end{figure*}

\section{Problem Formulation}
We formalize the genomic selection problem as a mapping learning task. Given a genotype-phenotype pair $(x, y)$, the final goal is to learn a mapping (or function) $f: x \mapsto y$.
In this work, the genotype data $x$ is the raw SNP sequence. Let $S=\{s_{1},s_{2},\dots,s_{N_l}\}$ denote the SNP sequence, where $N_l$ is the sequence length. We have $s_{i}\in\{\texttt{A},\texttt{T},\texttt{C},\texttt{G},\texttt{N},\texttt{Y},\texttt{K},\texttt{W},\texttt{R},\texttt{S},\texttt{M}\}$, where different letters represent different combinations of alleles~\cite{johnson2010extended}. In this work, the definition of each letter is presented in Tab.~\ref{tab:def}.
The phenotype data $y$ represents the corresponding phenotype values, which can either be discrete or continuous, depending on the task of interest (\ie~classification or regression). 

Let $N_s$ denote the number of samples for the species of interest in a crop dataset. In contrast to common machine learning tasks, we have $N_s \ll N_l$. This poses a non-trivial challenge for overfitting-prone methods such as Transformer, and thus Transformer cannot be directly used.

\section{Methodology}                                        
The focus of our framework is to mitigate the limitations of Transformer and capture the contextual information and language patterns of SNP sequence. To this end, we first pre-process raw SNP sequence by an initial mapping which does not change its length. After that, we implement k-mer tokenization and random masking to convert SNP sequence to token ID sequence. Then, embedding vectors are generated by the embedding layer and positional encoding~\cite{vaswani2017attention}. Finally, the embedding vectors are mapped to phenotype predictions by a neural network. The overall end-to-end training architecture is elaborated in Fig.~\ref{fig:System Model}.

\subsection{Pre-processed SNP Sequence}
In the candidate list of the sequence, $\{\texttt{A},\texttt{T},\texttt{C},\texttt{G}\}$ is the majority and $\{\texttt{N},\texttt{Y},\texttt{K},\texttt{W},\texttt{R},\texttt{S},\texttt{M}\}$ is the minority. To avoid the difficulty of excessively large k-mer vocabularies contributed by the minority and improve the computational efficiency of subsequent self-attention, we generate the pre-processed SNP sequence $S^{p}$ by the mapping rule:
\begin{equation}
\begin{aligned}
    s_{i}^p =\left\{
    \begin{array}{ll}
        s_{i}, & s_{i}\in\{\texttt{A},\texttt{T},\texttt{C},\texttt{G}\} \\
        \texttt{X}, & s_{i} \notin \{\texttt{A},\texttt{T},\texttt{C},\texttt{G}\}
    \end{array}
\right.
\end{aligned}
\end{equation}

\subsection{Sequence Tokenizer}
Considering the biological significance of SNP sequences and mitigating the attention computational cost on long sequences, we choose non-overlapping k-mer as a tokenization technique on pre-processed SNP sequence $S^{p}$. Let $T= \{\dots,\underline{t_{i}},\underline{t_{i+1}},\underline{t_{i+2}},\dots\}$ denote the token ID sequence processed by k-mer operation. Here is an illustration for a 4-mer operation with simulated sequence and token IDs. 
\begin{equation}
\begin{aligned}
T &= mer(S^{p}, k=4)\\
&=\{\dots,\underline{\texttt{A},\texttt{C},\texttt{G},\texttt{T}},\underline{\texttt{X},\texttt{A},\texttt{A},\texttt{C}},\underline{\texttt{G},\texttt{T},\texttt{X},\texttt{A}},\dots\}\\
&=\{\dots,\underline{39},\underline{502},\underline{346},\dots\}\in\mathbb{R}^{1\times \lfloor\frac{N_l}{k}\rfloor}
\end{aligned}
\nonumber
\end{equation}
$mer = (\cdot, k)$ is a function that treats the substring of length $k$ as a whole entity and converts it to a unique ID. Although k-mer shortens the SNP sequence to $1/k$ of original length $N_l$, the problem of $N_s \ll N_l$ still exists.

We introduce randomness by \textit{random masking} on token ID sequence to reduce the risk of overfitting:   
\begin{equation}
\begin{aligned}
    T^{m}= Mask(T)=\{\dots,t_{i}^m,t_{i+1}^m,t_{i+2}^m,\dots\}.
\end{aligned}
\end{equation}
$Mask(\cdot)$ is a function that has the pre-defined probability $p$ to convert the original token ID to $mask\_id$, which is defined separately. For the $i^\mathrm{th}$ position in $T^{m}$, we have
\begin{equation}
    t_{i}^m=Mask(t_{i})=
    \begin{cases}
      t_{i}, & p\\
    mask\_id, & 1-p
    \end{cases}   
\end{equation}

The final embedding vector is the sum of the embedded masked token ID sequence and the positional encoding.
\begin{equation}
    E_{0} = Embed(T_{m})+ E_{pos} 
\end{equation}
$Embed(\cdot)$ denotes the embedding layer, where the vocabulary size is $5^{k}+1$. The ``$5^{k}$'' represents the number of all possible combinations and ``$1$'' denotes the specific Mask ID for unknown cases. $E_{pos} \in\mathbb{R}^{1\times\lfloor\frac{N_{l}}{k}\rfloor\times D_{w}}$ is the sinusoidal positional encoding~\cite{vaswani2017attention}, where $D_{w}$ denotes embedding width.

\subsection{Phenotype Learning}
We use Transformer encoder to learn the contextual information of encoded SNP sequences, which consists of $L$ layers of multi-head self-attention (MSA) \cite{vaswani2017attention} and multi-layer perceptron (MLP) blocks.
\begin{gather}
E'_{l}=MSA(LN(E_{l-1}))+E_{l-1}\\
E_{l}=MLP(LN(E'_{l}))+E'_{l}
\end{gather}
$LN(\cdot)$ is the layer normalization~\cite{vaswani2017attention}.

After computing the attention, a linear projector $LP(\cdot)$ is used to map the output to a lower-dimensional space, thus lowering the computational complexity in the regressor:
\begin{equation}
    E_{P} = LP(E_{L})
\end{equation}

Finally, the phenotype prediction $y_{p}$ can be output as:
\begin{gather}
y_{p} = MLP(Flatten(E_{P})),
\end{gather}
where the projected vector is flattened and inputted into MLP layer to predict the phenotype.

There are two types of learning tasks in predicting phenotype from genotype information: variety recognition and trait prediction, corresponding to classification and regression tasks, respectively. 
For the classification task, we use the cross entropy (CE) as the loss function:
\begin{gather}
\mathcal{L}_{cls} = CE(y_{p}, y_{l}),
\end{gather}
where $y_{l}$ denotes the label.

For the regression task, we define the loss function by Mean Squared Error (MSE):
\begin{gather}
\mathcal{L}_{reg} = MSE({y_p}, y_{l}).
\end{gather}

\section{Experiments}
\subsection{Setup}
\paragraph{Datasets}
We use two crop datasets: \textit{rice3k}~\cite{wang2018genomic} and \textit{wheat3k}, representing the two most significant staple crops worldwide. Each dataset includes raw SNP sequences and multiple traits of interest (phenotype).

\textit{rice3k} is a public dataset that comprises genomic data and the corresponding phenotype data for a total of 3024 samples collected from 89 countries. For genomic data, the SNP sequence for each sample has a length of 404388. \textit{rice3k} encompasses multiple quality traits, \ie~discrete variables. We select six valuable and well-structured phenotypes for experimentation, including APCO\_REV\_REPRO, CUST\_REPRO, LPCO\_REV\_POST, LSEN, PEX\_REPRO, and PTH. The definitions of these phenotypes are described in Tab.~\ref{tab:rice3k_desc}.

\textit{wheat3k} is a private dataset to be released. \textit{wheat3k} is collected following the setup of \textit{rice3k} to provide a comprehensive study on wheat. It comprises genomic data and the corresponding phenotype data for a total of 3032 samples. Each SNP sequence for each sample has a length of 201740. In contrast to \textit{rice3k}, \textit{wheat3k} focuses on quantitative traits, \ie~continuous variables. We select six valuable and well-structured phenotypes for experimentation, including COLD, FHB, LODGING, STERILSPIKE, TKW, and YIELD. The definitions of these phenotypes are described in Tab.~\ref{tab:wheat_desc}.

\begin{table}[t]
\centering
\resizebox{\linewidth}{!}{
\begin{tabular}{l|c}\toprule
\textbf{Phenotype} & \textbf{Description} \\ \midrule
APCO\_REV\_REPRO & Apiculus color at reproductive  \\ 
CUST\_REPRO & Culm strength at reproductive - cultivated \\
LPCO\_REV\_POST & Lemma and palea color at post-harvest \\
LSEN & Leaf senescence \\
PEX\_REPRO & Panicle exsertion at reproductive \\
PTH & Panicle threshability \\
\bottomrule
\end{tabular}} 
\caption{Description of each phenotype of interest in the \textit{rice3k} dataset.}
\label{tab:rice3k_desc}
\end{table}

\begin{table}[t]
\centering
\resizebox{0.83\linewidth}{!}{
\begin{tabular}{l|c}\toprule
\textbf{Phenotype} & \textbf{Description} \\ \midrule
COLD & Resistance score to cold  \\ 
FHB & Resistance score to red rot disease \\
LODGING & Lodging severity \\
STERILSPIKE & Number of sterile spikes \\
TKW & Thousand kernel weight \\
YIELD & Wheat yield \\ \bottomrule
\end{tabular}}
\caption{Description of each phenotype of interest in the \textit{wheat3k} dataset.}
\label{tab:wheat_desc}
\end{table}

\paragraph{Implementation}
We implement the proposed framework using PyTorch and conduct experiments on NVIDIA 3090 GPU. There are a total 80 of epochs with early stops. The dimension of the sequence embedding width after embedding layer is set as 32 (\ie, $D_{w}=32$). Note that the vocabulary size of embedding layer is $5^{k}+1$, where 1 additional ID for unknown words. Following the settings in DNABERT \cite{ji2021dnabert}, we use 6-mer (\ie, $k=6$) and the default random masking ratio is $15\%$ (\ie, $p=0.15$). We use 3 Transformer encoder blocks in the learning module and an Adam optimizer~\cite{kingma2014adam} with an initial learning rate of 0.0001 and weight decay of 0.01. All experiments are repeated for five-fold cross-validation, following the standard practice in GS. The reported numbers are the average metrics and standard deviations. Following~\cite{wang2023dnngp,wu2024transformer}, each phenotype of interest is considered an independent task for training.

\paragraph{Evaluation Metrics}
Following the setups of~\cite{wang2023dnngp,wu2024transformer}, we adopt accuracy (ACC) as the evaluation metric for the classification task (\textit{rice3k} dataset) and Pearson correlation coefficient (PCC) to evaluate the regression performance (\textit{wheat3k} dataset) of our model. PCC is defined as.
\begin{equation}
    PCC = \frac{\sum_{i=1}^{n} (y_{p,i} - \bar{y}_{p})(y_{l,i} - \bar{y}_{l})}{\sqrt{\sum_{i=1}^{n} (y_{p.i} - \bar{y}_{p})^2 \sum_{i=1}^{n} (y_{l,i} - \bar{y}_{l})^2}}.
\end{equation}

\paragraph{Baselines}
To demonstrate the robustness of our method, we select four seminal methods for comparison, including GBLUP~\cite{vanraden2008efficient,yin2023hiblup}, RidgeClassifier~\cite{quyet2024mapping}, BayesA~\cite{meuwissen2001prediction,yin2022hibayes}, DNNGP~\cite{wang2023dnngp}, and DLGWAS~\cite{liu2019phenotype}.
GBLUP is a representative BLUP-based method and performs well across various regression tasks. To suit the classification tasks of \textit{rice3k}, we replace GBLUP with RidgeClassifier, which shares similar statistical assumptions. BayesA is a robust Bayesian method with simple assumptions. We conduct 1500 iterations to ensure convergence. DNNGP and DLGWAS are two state-of-the-art deep learning methods. We use the source code of DNNGP\footnote{\url{https://github.com/AIBreeding/DNNGP}}. We pre-process SNP sequences using PCA and retain 3000 principal components to preserve $95\%$ variance information. We implement DLGWAS as an integration of GWAS and Transformer. GWAS uses the \emph{mixed linear model}~\cite{muhammad2021uncovering} to analyze the correlation of SNP loci, retaining the top 10\% of loci based on effect size. After locus screening, the relative order of loci retained in the original sequence is preserved for Transformer modeling.
The model configuration information is kept as consistent as possible for all baselines. 

\begin{table*}[t]
\centering
\resizebox{0.97\linewidth}{!}{
\begin{tabular}{ccccccc}
\toprule
Method & APCO\_REC\_REPRO  & CUST\_REPRO   & LPCO\_REV\_POST& LSEN & PEX\_REPRO & PTH\\ \midrule
RidgeClassifier & 0.4872$\pm$0.0116&  0.4472$\pm$0.0192 & \textbf{0.5930$\pm$0.0163}   &  0.4858$\pm$0.0120& 0.5820$\pm$0.0145& {0.5106$\pm$0.0237}\\ 
BayesA &  0.4597$\pm$0.0154& 0.3521$\pm$0.0169& 0.5113$\pm$0.0176& 0.4133$\pm$0.0132&  0.5764$\pm$0.0101& 0.4975$\pm$0.0178\\ 
DNNGP & 0.2808$\pm$0.0166& 0.2441$\pm$0.0146&  0.2755$\pm$0.0121& 0.2272$\pm$0.0108& 0.1430$\pm$0.0174& 0.1675$\pm$0.0258\\ 
DLGWAS&  0.4863$\pm$0.0154 & 0.4538$\pm$0.0126  &  0.5826$\pm$0.0109 & 0.4955$\pm$0.0126 & 0.6139$\pm$0.0059 & 0.5346$\pm$0.0266\\ 
Ours & \textbf{0.5018$\pm$0.0180}  & \textbf{0.4685$\pm$0.0135} & {0.5922$\pm$0.0103} & \textbf{0.5058$\pm$0.0108} & \textbf{0.6189$\pm$0.0059} & \textbf{0.5422$\pm$0.0242}   \\
\bottomrule
\end{tabular}}
\caption{Comparison with state-of-the-art methods on the \textit{rice3k} dataset. On average, our method is 2.06\% superior to the best-performing statistical baseline RidgeClassifier and 1.05\% superior to the best-performing deep learning baseline DLGWAS on average across 6 traits.}
\label{tab:rice3k_result}
\end{table*}

\begin{table*}[h]
\centering
\resizebox{0.97\linewidth}{!}{
\begin{tabular}{ccccccc}\toprule
{Method} & {COLD} &{FHB} & {LODGING}  & {STERILSPIKE} & {TKW} & YIELD\\ \midrule
GBLUP &0.3587$\pm$0.0324 &0.2447$\pm$0.0486 &0.6195$\pm$0.0178 & 0.3983$\pm$0.0359 & 0.5676$\pm$0.0204 & 0.5896$\pm$0.0355\\ 
BayesA &0.4691$\pm$0.0194 &0.3027$\pm$0.0451 &0.8041$\pm$0.0138 & 0.6813$\pm$0.0215 & 0.7462$\pm$0.2147 &  0.6968$\pm$0.0278\\ 
DNNGP &0.2687$\pm$0.0282 &0.0122$\pm$0.0427 &0.5029$\pm$0.0200 & 0.3089$\pm$0.0304 & 0.5698$\pm$0.2315 & 0.0926$\pm$0.0192\\ 
DLGWAS & 0.4781$\pm$0.0085 & 0.3354$\pm$0.0524 & 0.8166$\pm$0.0203 & 0.6794$\pm$0.0154 & 0.7395$\pm$0.0255 & 0.7000$\pm$0.0223 \\ 
Ours & \textbf{0.4937$\pm$0.0120} & \textbf{0.3408$\pm$0.0534} & \textbf{0.8270$\pm$0.0136} & \textbf{0.6862$\pm$0.0175} & \textbf{0.7541$\pm$0.0213} & \textbf{0.7163$\pm$0.0159}  \\ 
\bottomrule
\end{tabular}}
\caption{Comparison with state-of-the-art methods on the \textit{wheat3k} dataset. On average, our method is 1.97\% superior to the best-performing statistical baseline BayesA and 1.15\% superior to the best-performing deep learning baseline DLGWAS on average across 6 traits.}
\label{tab:wheat_result}
\end{table*}

\subsection{Results}
\paragraph{Performance on Classification Tasks}
Tab.~\ref{tab:rice3k_result} elaborates the classification results of different models on \textit{rice3k} dataset. On the phenotypes of APCO\_REV\_REPRO, CUST\_REPRO, LPCO\_REV\_POST, LSEN, PEX\_REPRO, and PTH, our model achieves the best performance compared to the existing models in ACC metric. On LPCO\_REV\_POST, we also obtain competitive results. Specifically, we surpass the current best-performing method DLGWAS over 1.05\% across these 6 traits on average. The robustness and advancement of our simple approach to enhancing Transformer are evidenced by the overall elevation in mean values and reduction in standard deviation. These outcomes position our method as a pioneering benchmark in the field. 

\paragraph{Performance on Regression Tasks}
We further analyze each baseline's capacity on \textit{wheat3k} dataset, tabulated in Tab.~\ref{tab:wheat_result}. Similar to the outstanding performance on the \textit{rice3k} dataset, our model also achieves the best performance compared to existing methods, where the higher PCC metric and lower standard deviation clearly prove the superiority. On YIELD task, which is important as it signifies the meaning of wheat yield, our method is 1.63\% superior to the current state-of-the-art DLGWAS, which holds the potential in the application of genomic selection for crop breeding.

\subsection{Ablation Studies}
To investigate the contribution of each key component of our proposed method, we conduct a series of ablation experiments on the \textit{wheat3k} dataset.

\begin{table}[t]
\centering
\resizebox{\linewidth}{!}{
\begin{tabular}{ccccc}\toprule
\multicolumn{2}{c}{Method} & FHB & STERILSPIKE & YIELD \\ \midrule
\multicolumn{2}{c}{DLGWAS} & 0.3354$\pm$0.0524& 0.6794$\pm$0.0154& 0.7000$\pm$0.0223\\ \midrule
{k-mer} & {Random Masking} & & & \\
\cmidrule(r){1-2}
 &     & 0.3216$\pm$0.0360 & 0.5881$\pm$0.0098 & 0.6663$\pm$0.0201 \\ 
 & \Checkmark & 0.3290$\pm$0.0406 & 0.5921$\pm$0.0090 & 0.6699$\pm$0.0205  \\ 
 \Checkmark& &  0.3340$\pm$0.0539  & 0.6723$\pm$0.0179   & 0.7044$\pm$0.0123 \\ 
 \Checkmark& \Checkmark & \textbf{0.3408$\pm$0.0534} & \textbf{0.6862$\pm$0.0175} & \textbf{0.7163$\pm$0.0159} \\ 

\bottomrule
\end{tabular}}
\caption{Ablation study of the different component combinations on the \textit{wheat3k} dataset. In this table, we use 6-mer, and random masking proportion is set as 15\%. We find both components are beneficial to enhance model capacity, while k-mer is more important.}
\label{tab:component}
\end{table}

\begin{figure}[t]
        \centering
        \includegraphics[width=0.97\linewidth]{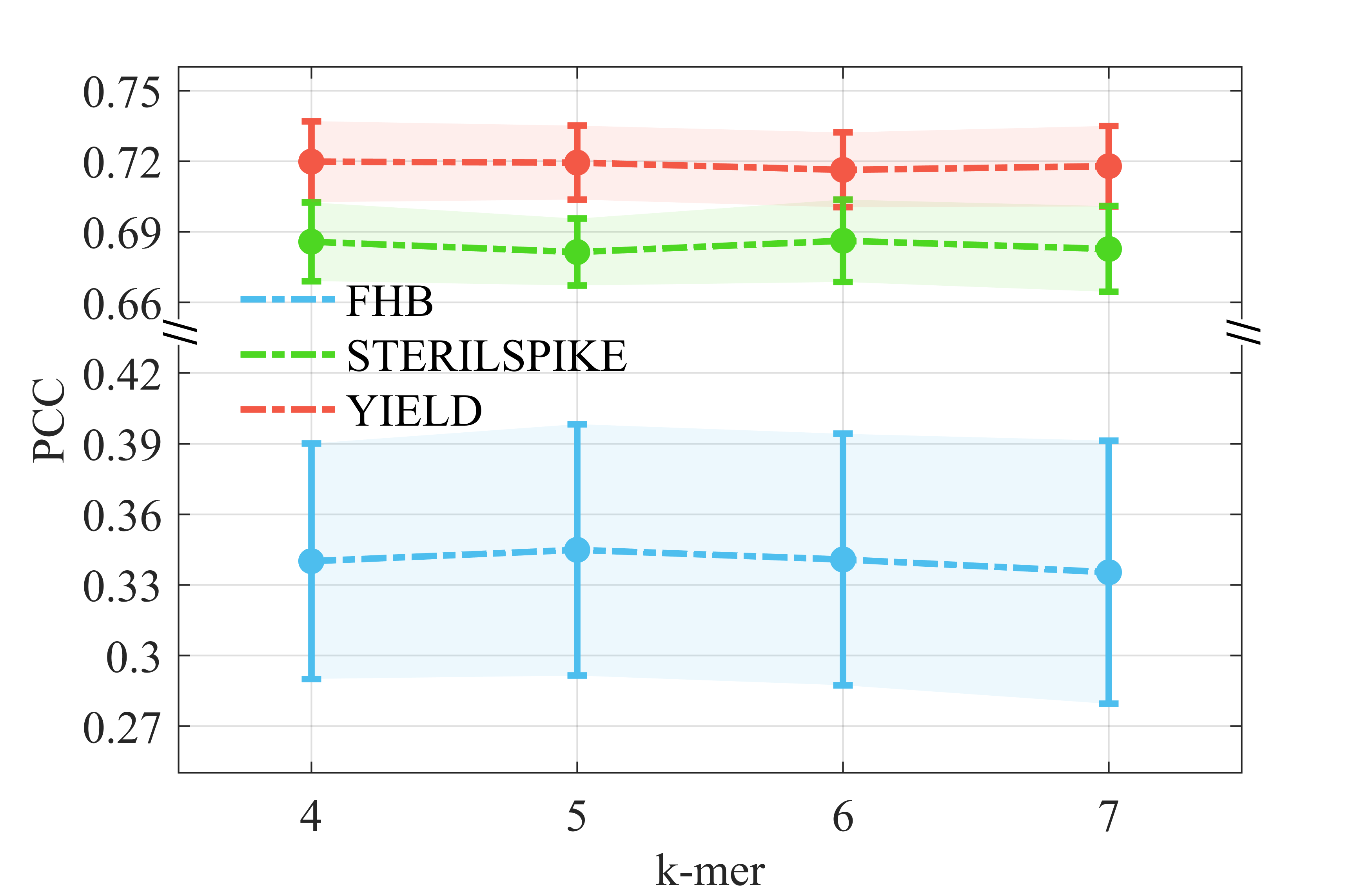}
    \vspace*{-2mm}
    \caption{Sensitivity analysis on $k$ for k-mer tokenization on the \textit{wheat3k} dataset. The optimal value of $k$ might be dependent on the task (phenotype).}
    \label{ablation_kmer}
\end{figure}  

\begin{table*}[t]
\centering
\resizebox{0.93\linewidth}{!}{
\begin{tabular}{ccccc}\toprule
{Method}  & Vocab Size &  {FHB}  & {STERILSPIKE} & {YIELD}\\
\midrule
{Character-level} &  5\textsuperscript{\phantom{0}}+ 1 (for unknown) & {0.3216$\pm$0.0360}   & {0.5881$\pm$0.0098}   & {0.6663$\pm$0.0201} \\ 
Byte-Pair Encoding & 8000  & 0.0562$\pm$0.0123  & 0.3159$\pm$0.0341 &  0.3656$\pm$0.0447  \\ 
 Learnable Tokenizer (MLP)&  {-} & {0.3092$\pm$0.0576}  & {0.6669$\pm$0.0307}   & {0.6747$\pm$0.0590}  \\ 
Learnable Tokenizer (Transformer)& {-} & {0.3329$\pm$0.0560}  & {0.6592$\pm$0.0219}   & {0.7086$\pm$0.0166}  \\ 
 {Ours (5-mer+15\% random masking)} & 5\textsuperscript{5}+ 1 (for unknown)& {\textbf{0.3449$\pm$0.0534}}   & {\textbf{0.6814$\pm$0.0142}} & {\textbf{0.7194$\pm$0.0157}}  \\ 
\bottomrule
\end{tabular}}
\caption{Comparison of different tokenizer methods on the \textit{wheat3k} dataset. In character-level tokenizer and our method, there exists an additional token in vocabulary for unknown cases. Under 5-mer and 15\% random masking, our model presents 1.20\%, 2.22\%, and 1.08\% improvements over the state-of-the-art tokenizers on FHB, STERILSPIKE, and YIELD, respectively. In this table, we set $k=5$ in k-mer for our model in comparison as it holds the overall best performance in Fig.~\ref{ablation_kmer}.}
\label{tokenizer}
\end{table*}

\paragraph{Analysis of Pre-Processing}
We conduct ablation experiments to have a deep insight into key components in the tokenization. Note, we choose the model without pre-processing module as baseline to study the impact of each component in Tab.~\ref{tab:component}. It is found that k-mer plays a more significant role than random masking, which enhances the performance of baseline model by 1.24\%, 8.42\%, and 3.81\% in PCC on three sub-tasks of \textit{wheat3k} dataset. This suggests that using k-mer in pre-processing can assist attention mechanisms to better capture contextual information. Another finding is that when utilizing k-mer as tokenizer, random masking can yield greater improvements than using random masking on baseline alone. The difference lies in whether the smallest unit of the masking is 1 bit or $k$ bits, which relates to the influence of masking on context. This proves the importance of context extraction on k-mer. Compared with DLGWAS, Transformer without k-mer and random masking shows inferior performance and our model enhances the performance. This proves the validity of key components.

\paragraph{Analysis of k-mer Tokenization}
We study the influence of $k$ in k-mer tokenizer in Fig.~\ref{ablation_kmer}. 
Note, when $k=5$, our model obtains the best results on the FHB and YIELD traits, but the best results on the STERILSPIKE occur at $k=6$. We hypothesize that the value of $k$ might be task-oriented.

\paragraph{Analysis on Random Masking}
\begin{table}[t]
\centering
\resizebox{\linewidth}{!}{
\begin{tabular}{cccc}\toprule
{Proportion}  &{FHB}  & {STERILSPIKE} & {YIELD}\\
\midrule
0\% &   0.3340$\pm$0.0539  & 0.6723$\pm$0.0179   & 0.7044$\pm$0.0123 \\ 
 15\%& 0.3408$\pm$0.0534 & 0.6862$\pm$0.0175 & 0.7163$\pm$0.0159 \\ 
 30\%&  \textbf{0.3456$\pm$0.0509}  & \textbf{0.6883$\pm$0.0188}   &  \textbf{0.7169$\pm$0.0134} \\ 
 45\% & 0.3394$\pm$0.0533  & 0.6861$\pm$0.0170  &  0.7147$\pm$0.0165 \\ 
  60\% & 0.3380$\pm$0.0508   & 0.6738$\pm$0.0204 & 0.7093$\pm$0.0170 \\ 
\bottomrule
\end{tabular}}
\caption{Impact of the masking proportion for the capacity of our method on the \textit{wheat3k} dataset. When 30\% tokens are masked for a SNP sequence, our model achieves the best performance.}
\label{random mask}
\end{table}

We conduct experiments on random masking proportion to study the effect of randomness in the tokenization. In Tab.~\ref{random mask}, the PCC value of three sub-tasks improves as the random masking proportion increases, reaching its peak value when the proportion equals to 30\%. Through randomly masking tokens, the data diversity is enhanced, thereby promoting the model's generalization ability and alleviating overfitting issues arising from data scarcity, especially in our problem formulation ($N_s \ll N_l$). When the masking ratio becomes too high, \ie, 45\%, it can lead to diminishing returns and worsen the effectiveness of the model because excessively masking a large portion of tokens may disrupt the coherence of the input sequence to make it harder for the model to learn meaningful representations.

\paragraph{Choice of Tokenizer}
We compare our method with seminal tokenization algorithms to demonstrate the effectiveness of k-mer and random masking, shown in Tab.~\ref{tokenizer}. Character-level tokenizer treats each letter in SNP sequence as an independent feature, where we use $\{\texttt{A},\texttt{T},\texttt{C},\texttt{G},\texttt{X}\}$ and one additional ID for unknown words as the vocabulary. Byte-Pair Encoding builds a sub-word vocabulary of size 8000.  We also implement two learnable tokenizers: one MLP layer and one Transformer encoder, where the input is SNP sequence and the embedding width of encoding output is 20. Compared with the second-best tokenizer (Learnable Transformer), our method achieves 1.20\%, 2.22\%, and 1.08\% PCC improvements on FHB, STERILSPIKE, and YIELD, respectively, providing better SNP sequence representation for learning module. 

\begin{table}[t]
\centering
\resizebox{\linewidth}{!}{
\begin{tabular}{cccc}\toprule
{Method} & {Parameters (M)} &  {Memory (MB)}  & {Time (ms)}\\
\midrule
  GBLUP  &  -   & - &   \phantom{00}11.00\\
  BayesA  &  -   & - &   1143.00\\ 
 DNNGP  &  \phantom{00}0.35 & \phantom{000}0.36  &   \phantom{000}0.40\\ 
 DLGWAS  & \phantom{0}34.54  & \phantom{0}185.31  &  \phantom{00}27.10  \\ 
 Transformer & 206.69 & 1514.94  & \phantom{0}899.22 \\
  Ours (5-mer)  & \phantom{0}41.52 & \phantom{0}246.16  & \phantom{00}39.29  \\ 
  Ours (6-mer)  & \phantom{0}35.04 & \phantom{0}205.68  & \phantom{00}26.91  \\ 
\bottomrule
\end{tabular}}
\caption{Computational cost of the state-of-the-art methods on \textit{wheat3k} dataset. Mentioned that our computational cost are statistically based on individual samples. Compared with the statistical method BayesA and \emph{vanilla} Transformer (without tokenization), our method successfully reduces computation cost.}
\label{tab:calculation}
\end{table}

\paragraph{Computational Cost}
Lastly, to comprehensively assess a model, computational cost is considered. Take \textit{wheat3k} as an example, the original sequence length is 201740. Tabulated in Tab.~\ref{tab:calculation}, we utilize parameter count (Parameters), GPU memory consumption (Memory), and inference time (Time) to measure all models. GBLUP shows fast inference time as a lightweight linear model.  But it sacrifices the degree of freedom, thus leading to lower accuracy in Tab.~\ref{tab:wheat_result}. After PCA, DNNGP only retains 3000 principal components for CNN, thus generating low computational costs. However, DNNGP reports the lowest performance on both two datasets. For \emph{vanilla} Transformer without tokenization, the original SNP sequence of length 201740 is input into attention mechanism. For DLGWAS and Ours (6-mer), the inputs for both Transformer encoder consist of 33623 dimensions. For Ours (5-mer), this value is 40348. 
Compared with the statistical method BayesA, our method provides a quicker inference time, and compared with \emph{vanilla} Transformer, our method efficiently reduces GPU memory cost.  

\section{Conclusion}
In this study, we present a simple yet effective approach to enhance Transformer's performance in the realm of genomic selection for crop breeding. Specifically, we propose the pre-processing module composed of k-mer tokenizer and random masking to assist contextual understanding of SNP sequence. Experiments on the \textit{rice3k} and \textit{wheat3k} datasets demonstrate promising performance on genotype-to-phenotype prediction. The empirical findings of this work not only suggest the potential of DL to handle long sequences but also pose a new research direction on end-to-end genomic selection.

Meanwhile, we shall notice that though tokenization plays an important role in genomic selection, a comprehensive understanding is still needed in future work.

\clearpage
\section*{Acknowledgements}
This work was supported by Shanghai Artificial Intelligence Laboratory and the Strategic Priority Research Program of the Chinese Academy of Sciences under Grant No.XDA0450203.
\section*{Contribution Statement}
Renqi Chen, Wenwei Han, and Haohao Zhang have equal contributions. Nanqing Dong is the corresponding author.

\bibliographystyle{named}
\bibliography{ijcai24}
\end{document}